# SEMI-PointRend: Improved Semiconductor Wafer Defect Classification and Segmentation as Rendering

MinJin Hwang[a,b*], Bappaditya Dey[a*], Enrique Dehaerne[a,c*], Sandip Halder[a], Young-han Shin[b]

[a]imec, Kapeldreef 75, 3001 Leuven, Belgium; [b]Dept. of Physics, University of Ulsan, Ulsan, South Korea; [c]Dept. of Computer Science, KU Leuven, Leuven, Belgium

*These authors contributed equally

## ABSTRACT

In this study, we applied the PointRend (Point-based Rendering) method to semiconductor defect segmentation. PointRend is an iterative segmentation algorithm inspired by image rendering in computer graphics, a new image segmentation method that can generate high-resolution segmentation masks. It can also be flexibly integrated into common instance segmentation meta-architecture such as Mask-RCNN and semantic meta-architecture such as FCN. We implemented a model, termed as SEMI-PointRend, to generate precise segmentation masks by applying the PointRend neural network module. In this paper, we focus on comparing the defect segmentation predictions of SEMI-PointRend and Mask-RCNN for various defect types (line-collapse, single bridge, thin bridge, multi bridge non-horizontal). We show that SEMI-PointRend outperforms Mask R-CNN by up to 18.8% in terms of segmentation mean average precision.

**Keywords:** semiconductor defect inspection, metrology, lithography, stochastic defects, supervised learning, deep learning, defect classification, defect localization, defect mask segmentation, mask r-cnn, pointrend

## 1. INTRODUCTION

Continuous scaling down of patterning dimensions and scaling up of chip demand are driving the need for innovation in semiconductor metrology and inspection. Automatic defect inspection that is accurate and reliable is key to reduce engineering time and manufacturing cost. Recently, Convolutional Neural Network (CNN)-based approaches have shown promising results in defect localization and classification[1-4]. Most of these previous works either only classify an image as containing a defect of a certain type or predict the location of a defect in the form of a minimal bounding-box, which we term as defect detection, which generally ignores the precise geometry of the defect pattern itself. Dey et al.[5] used the Mask R-CNN[6] neural network architecture to predict the exact pixels in an image which describe a defect pattern precisely. This task is also known as instance segmentation.

In this work, a PointRend model[7] is trained and evaluated on a dataset of line-space pattern images with at least one defect instance per image. This model is referred to as SEMI-PointRend in the rest of this paper. These defects are labeled as one of five different defect classes. It is shown that SEMI-PointRend model can segment defects more precisely than Mask R-CNN, outperforming it by up to 18.8% in terms of the segmentation mean Average Precision (mAP) metric. The implications of highly precise defect segmentation for future work include fine-grained and on-the-fly defect classification similarity measurements.

The rest of the paper is organized as follows. Section 2 discusses the work related to this work. Section 3 explains the proposed method based on the PointRend neural network module. The Scanning Electron Microscope (SEM) image dataset of semiconductor defects used to train and evaluate SEMI-PointRend as well as a baseline Mask R-CNN model is described in Section 4. Section 5 details the setup of experiments conducted. Section 6 shows results from these experiments and discusses them. Finally, Section 7 discusses limitations of this study and promising directions for future work.

## 2. RELATED WORK

In this section, previous research related to deep learning-based semiconductor defect segmentation is summarized. First, CNN-based semiconductor defect detection is summarized. Second, semiconductor defect segmentation is summarized.

Cheon et al.[1] proposed a method that used CNNs with k-Nearest Neighbors (k-NN) to extract features for wafer surface defect classification. The k-NN algorithm serves to classify new defect types that have not been seen during training as an *unseen* class. A limitation of this approach is that CNN feature vectors must be stored to calculate clusters of defects. Lei et al.[2] demonstrated that CNNs can achieve higher accuracies for defect classification compared to traditional (non-deep learning) models (etc. Random Forest, Support vector machine). They showed approximately 1.2% improvement over the best existing traditional model. Yuan-Fu and Min[3] proposed double features extraction method for wafer defect classification. They implemented the model that combined Radon transform operation with a CNN, called RadonNet. Radon transform is a technology that detects feature in an image. They showed that RadonNet can be used to classify 12 different types to defects. Dey et al.[4] noticed that certain CNN models can detect different defect classes better or worse than others. It was shown that combining predictions of these different models led to an improved mAP score.

Recently, Dey et al.[5] proposed defect classification and segmentation in SEM images using the Mask R-CNN[6] model architecture. Mask R-CNN is a model that adds a mask header network to Faster R-CNN[8] to enable mask output in addition to classification and bounding boxes. It was shown that Mask R-CNN can extract defect instance segmentation masks in addition to classifying and detecting defects in SEM images. Therefore, it can predict the location, class, shape, and surface area (in pixels) of defect instances. A limitation of Mask R-CNN is that its segmentation predictions were often not precise and included artifacts that were not part of a defect. In this study, the more recent PointRend[7] neural network module is proposed as the solution to this limitation since it can predict more fine-grained instance segmentations.

## 3. PROPOSED METHODOLOGY

In this, section, we propose a more accurate instance segmentation and precise mask generation method for semiconductor defect patterns using the PointRend[7] neural network module. The PointRend module is inspired by the adaptive subdivision[9] technique used primarily in the field of computer graphics. This module takes as input a coarse CNN feature map and labels fine points in this map through recursively up-sampling points of the feature map. During inference, the module selects a pre-defined number of points which are likely to be border points, up-samples them, and labels the up-sampled points. This procedure is repeated multiple times until sufficiently fine-grained points are inspected and labeled, as shown in Figure 1. This results in efficient fine-grained segmentation. During training, a point sampling method biased towards points closer to the border of segmentations is employed to avoid iterative up-sampling and labeling.

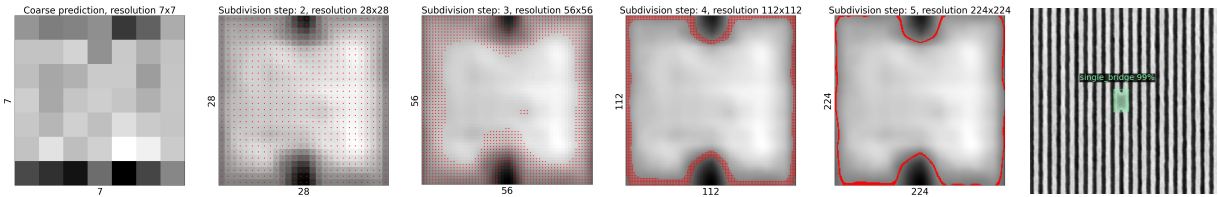

Figure 1. Point change according to the subdivision step of the proposed method as applied on a coarse feature map (most left image) of a single-bridge defect found in a SEM image (most right image).

The proposed model, referred to as SEMI-PointRend in the rest of this paper, consists of the Mask R-CNN model with its mask header replaced by the PointRend segmentation module and is trained on the dataset that will be introduced in Section 4. In this study, SEMI-PointRend and the original Mask R-CNN model are trained and evaluated on the same dataset to demonstrate that the PointRend module results in more precise segmentations compared to the segmentation header used by Mask R-CNN. An overview of the full SEMI-PointRend model is shown in Figure 2.

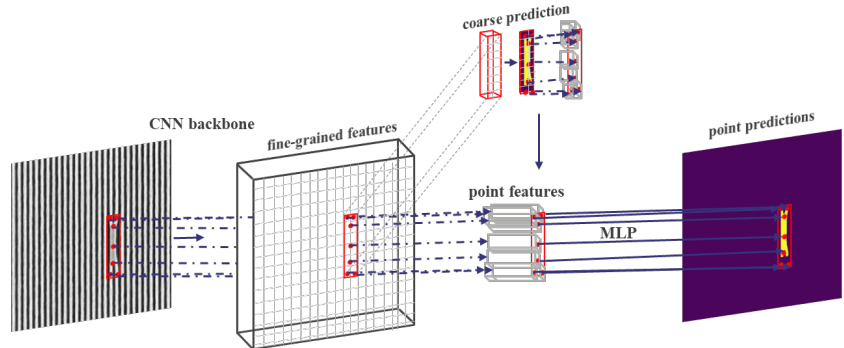

Figure 2. Overview of the SEMI-PointRend model.

## 4. DATASET

The dataset used to train and evaluate the proposed model, SEMI-PointRend, as well as the baseline model, Mask R-CNN, includes Scanning Electron Microscope (SEM) images of both post-litho and post-etch P32 (Pitch 32nm) line-space patterns with defects. The dataset consists of 1160 raw SEM images (480×480) pixels in TIFF format. The defects included in the dataset are classified as being either a thin bridge, single bridge, line collapse, multi bridge horizontal, or multi bridge non-horizontal. The labels for these defects were obtained using the VGG Image Annotator (VIA)[10]. We divided dataset for training, validating, and testing as shown in Table 1.

Table 1. Data distribution of defect images

| Class Name | Train | Val | Test |
|---|---|---|---|
| Thin bridge | 240 | 30 | 30 |
| Single bridge | 240 | 30 | 30 |
| multi bridge horizontal | 160 | 20 | 20 |
| multi bridge non-horizontal | 80 | 10 | 10 |
| Line collapse | 200 | 30 | 30 |
| Total | 920 | 120 | 120 |

## 5. EXPERIMENTAL SETUP

Both the SEMI-PointRend and Mask R-CNN models were initialized using the pre-trained ResNet101[11] backbone provided by Detectron2[12]. Hyperparameters used for training both models are as follows: 5000 epochs, image per batch size of 2, and learning rate of 0.00025. For both training and inference, an intersection-over-union (IOU) threshold of 0.5 was used. The metric used to evaluate the models is the Average Precision (AP) as calculated by COCO[13] each class and the mean over all classes (mAP) of bounding box or segmentation predictions compared to the expert annotations.

The experiments are implemented using Detectron2 and PyTorch[14] in a Python programming environment. Our model has been trained and evaluated on a Lambda Workstation with two NVIDIA RTX 3070 GPUs.

## 6. RESULTS & DISCUSSION

In this section, we compare the test results between the proposed SEMI-PointRend model and the baseline Mask R-CNN model. Table 2 provides bounding box AP and segmentation AP of each class for each model. The per-class segmentation APs achieved by SEMI-PointRend for all defect classes are always better or at least as good as those achieved by Mask R-CNN. The bounding box and segmentation mAPs achieved by the SEMI-PointRend model are 11.8% and 13.8% improved, respectively, compared to Mask R-CNN. This shows that SEMI-PointRend can generate masks more precisely than Mask R-CNN for every defect class.

Table 2. Test results for both Mask R-CNN and SEMI-PointRend per class and overall (mean AP) averaged over predictions using IOU thresholds between 0.5 and 0.95 with a step size of 0.05.

| Class Name | Mask R-CNN | | SEMI-PointRend | |
|---|---|---|---|---|
| | BBox AP | Segmentation AP | BBox AP | Segmentation AP |
| Thin bridge | 0.526 | 0.525 | **0.588** | **0.550** |
| Single bridge | 0.692 | 0.700 | **0.765** | **0.777** |
| Multi bridge horizontal | 0.316 | 0.366 | **0.528** | **0.535** |
| Multi bridge non-horizontal | **0.792** | **0.574** | 0.696 | 0.574 |
| Line collapse | **0.687** | 0.604 | **0.687** | **0.636** |
| Total (mAP) | 0.584 | 0.542 | **0.653** | **0.617** |

Table 3 shows the mAP results for both models for various IOU thresholds and calculated according to each IOU threshold score and object size. The improvements in mAP are largest when using an IOU threshold of 0.75 and for defects with large areas (18.8% and 18.4%, respectively). This suggests that SEMI-PointRend's segmentations are indeed more precise, and this is most important when segmenting large defects which usually have very complex shapes. This can be seen in Figure 3, where Mask R-CNNs predictions for the multi bridge defects include pixels which are not part of the defect patterns. In comparison, Figure 4 shows that SEMI-PointRend segments the defect patterns much more precisely.

Table 3. Comparison analysis of defect segmentation when experimenting with Mask R-CNN vs SEMI-PointRend.

| mAP with... | Mask R-CNN | | SEMI-PointRend | |
|---|---|---|---|---|
| | Bounding Box | Segmentation | Bounding Box | Segmentation |
| IOU 0.5:0.95 | 0.584 | 0.542 | **0.653** | **0.617** |
| IOU 0.5 | 0.871 | 0.853 | **0.994** | **0.994** |
| IOU 0.75 | 0.655 | 0.605 | **0.763** | **0.719** |
| Medium area ($32^2$ < pixel area < $96^2$) | 0.600 | 0.574 | **0.671** | **0.642** |
| Large Area (pixel area > $96^2$) | **0.602** | 0.353 | 0.552 | **0.418** |

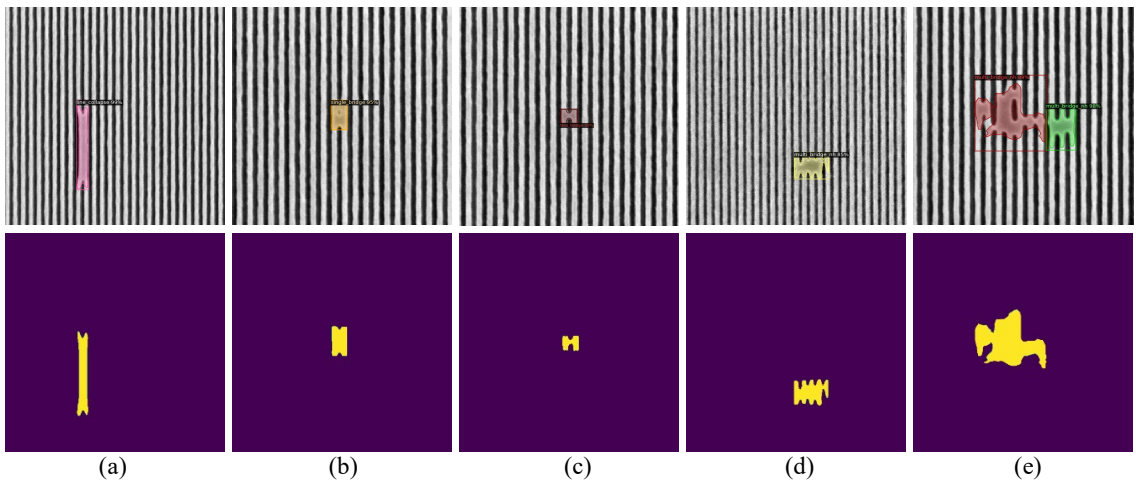

(a) (b) (c) (d) (e)

Figure 3. Example defect detection and segmentation results using Mask R-CNN for class (a) Line collapse, (b) Single bridge, (c) Thin bridge, (d) Multi bridge horizontal, and (e) Multi bridge non-horizontal. The top row shows SEM images with Mask R-CNN predictions while the bottom row shows corresponding binary images of predicted segmentations.

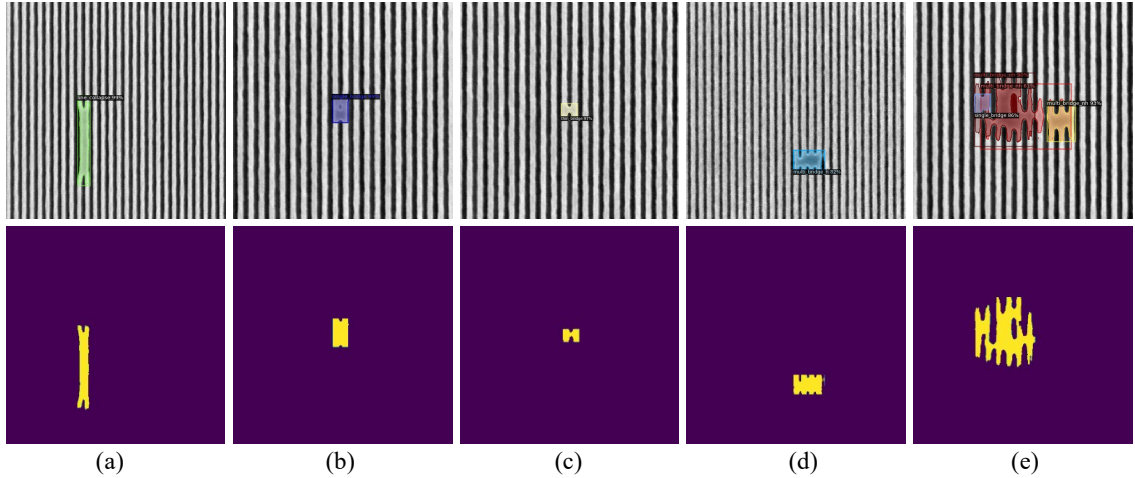

|     (a)     |     (b)     |     (c)     |     (d)     |     (e)     |

Figure 4. Example defect detection and segmentation results using SEMI-PointRend for class (a) Line collapse, (b) Single bridge, (c) Thin bridge, (d) Multi bridge horizontal, and (e) Multi bridge non-horizontal. The top row shows SEM images with SEMI-PointRend predictions while the bottom row shows corresponding binary images of predicted segmentations.

Precise segmentation enables accurate analysis of geometrical attributes of defect patterns such as area, perimeter, polygon shapes, arc lengths, etc. Tables 4 and 5 show defect mask area size statistics of the test set as obtained by analyzing predictions from Mask R-CNN and SEMI-PointRend, respectively. These statistics include the number of generated masks as well as the mean, standard deviation, minimum, and maximum of mask area for each defect class. Figure 5 visualizes these statistics using box plots. Figure 6 shows how these statistics (and others not used in our experiments, such as mask shape) can be used to analyze the (dis-)similarity between different defect patterns as well as between multiple instances of the same class.

Table 4. Mask area analysis statistics for Mask R-CNN.

| Defect Class | Count | Mean | Std | Min | 25% | 50% | 75% | Max |
|---|---|---|---|---|---|---|---|---|
| Thin bridge | 30 | 947.9 | 152.6 | 782.5 | 841.6 | 879.5 | 1020.6 | 1398.0 |
| Single bridge | 28 | 1090.1 | 358.3 | 693.0 | 828.0 | 982.8 | 1269.4 | 1962.5 |
| Multi bridge horizontal | 5 | 1351.7 | 205.5 | 1122.5 | 1244.0 | 1257.0 | 1531.5 | 1603.5 |
| Multi bridge non-horizontal | 33 | 2413.2 | 1713.0 | 1055.5 | 1389.0 | 1900.5 | 2881.5 | 10232.0 |
| Line collapse | 34 | 1676.8 | 377.5 | 1047.5 | 1419.8 | 1587.8 | 1858.8 | 2563.5 |

Table 5. Mask area analysis statistics for SEMI-PointRend.

| Defect Class | Count | Mean | Std | Min | 25% | 50% | 75% | Max |
|---|---|---|---|---|---|---|---|---|
| Thin bridge | 29 | 917.7 | 143.5 | 761.5 | 816.0 | 858.5 | 992.5 | 1277.5 |
| Single bridge | 34 | 1112.0 | 347.9 | 667.0 | 879.9 | 1027.0 | 1251.1 | 1979.5 |
| Multi bridge horizontal | 10 | 1502.9 | 429.7 | 1044.0 | 1219.6 | 1338.3 | 1617.0 | 2315.5 |
| Multi bridge non-horizontal | 25 | 3235.1 | 2832.9 | 1323.0 | 1663.0 | 2624.5 | 3220.0 | 12969.0 |
| Line collapse | 34 | 1739.7 | 385.4 | 1087.0 | 1488.5 | 1666.3 | 1921.5 | 2674.0 |

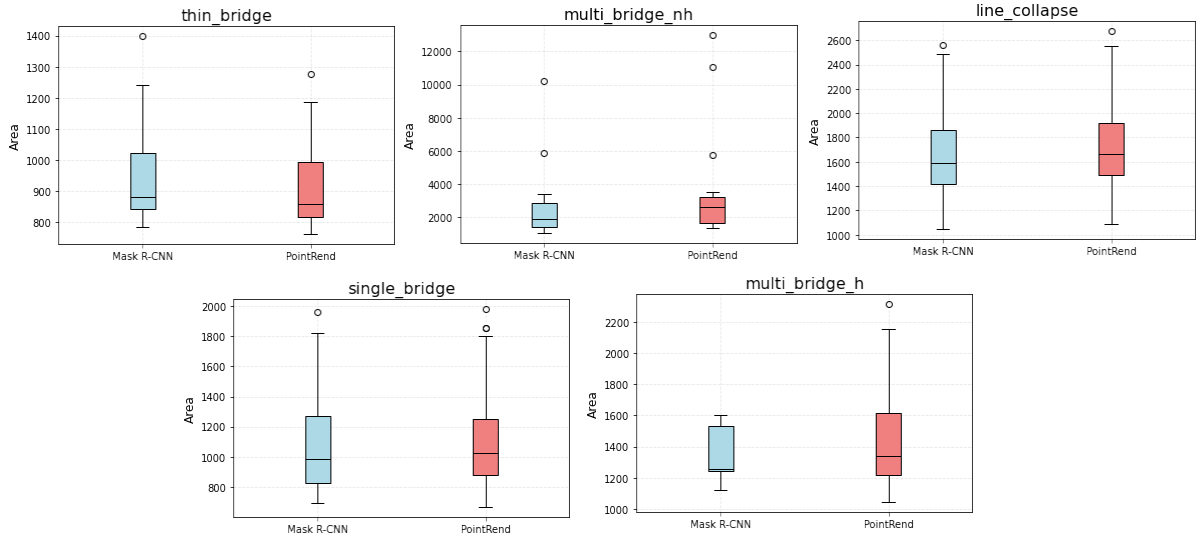

Figure 5. Boxplot of the predicted mask area size for each defect type and model.

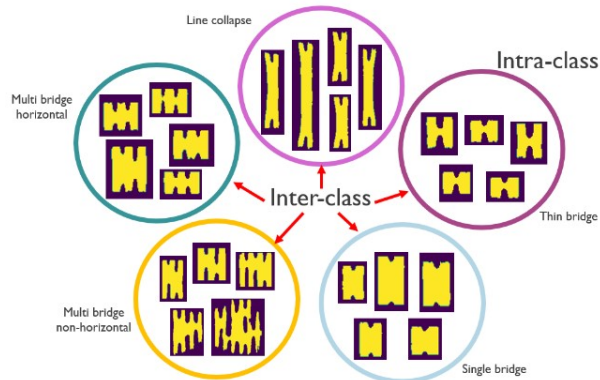

Figure 6. Statistics of predicted defect pattern segmentations, such as segmentation area, can be used to differentiate between different defect classes (inter-class) and between different defect instances of the same class (intra-class).

## 7. LIMITATIONS & FUTURE WORK

A limitation of supervised segmentation methods, especially very precise methods such as SEMI-PointRend, is that the learning is sensitive to the expert labeling of the dataset. The exact extent of a defect pattern can be ambiguous in certain cases, and this can confuse the model if different human experts annotate defects of the same class in different ways. Future work to address this issue could include employing data-centric approaches[15] or unsupervised learning.

A limitation of the experiments conducted in this study is that minimal hyperparameter tuning was performed. Finetuning general hyperparameters such as the learning rate or model specific hyperparameters such as the number of subdivision steps or points per subdivision step (see Section 3) would likely lead to improved results. Optimizing these and other hyperparameters specifically for defect segmentation in SEM images is suggested as future work.

## 8. CONCLUSION

In this paper, it was shown that SEMI-PointRend can segment semiconductor defects in SEM images more precisely than a standard Mask R-CNN model. The proposed model predicts segmentation masks efficiently using techniques inspired by rendering in computer graphics. Experimental results demonstrated that using SEMI-PointRend results in improving mAP of up to 18.8% compared to Mask R-CNN. The best improvements were shown to be achieved for high

IOU thresholds and large defects. The results suggest that SEMI-PointRend predictions can be used for more accurate and detailed analysis of defects in a dataset of semiconductor SEM images.

**Acknowledgements**: This research was supported by the MOTIE (Ministry of Trade, Industry, and Energy) in Korea, under the Fostering Global Talents for Innovative Growth Program (P0017312) supervised by the Korea Institute for Advancement of Technology (KIAT).